\documentclass[10pt, conference, compsocconf]{IEEEtran}
\IEEEoverridecommandlockouts

\usepackage[dvipsnames]{xcolor}

\usepackage{url}

\usepackage{graphicx,subcaption} 
\usepackage{booktabs}
\usepackage{amsfonts}

\usepackage{amsmath,amssymb,amsfonts}
\usepackage{graphicx}
\usepackage{textcomp}
\usepackage{xcolor}
\usepackage{balance}

\usepackage{multirow}

\usepackage[ruled,linesnumbered]{algorithm2e}
\usepackage{algorithmic}

\usepackage{pgfplots}
\pgfplotsset{compat=newest}
\pgfplotsset{plot coordinates/math parser=false}

\usepackage{cite}

\begin{document}

\title{Learn More by Using Less: Distributed Learning with Energy-Constrained Devices
}


\author{
\IEEEauthorblockN{Roberto Pereira,  Cristian J. Vaca-Rubio and Luis Blanco}
\IEEEauthorblockA{Centre Tecnol\`ogic de Telecomunicacions de Catalunya
(CTTC / CERCA)\\
email: \{rpereira, cvaca, lblanco\}@cttc.es
}
\thanks{
This work is supported by the SONATA project, funded by the grant CHIST-ERA-20-SICT-004 by PCI2021-122043-2A/AEI/10.13039/501100011033
and by the   SOFIA project PID2023-147305OB-C32, funded by MICIU/AEI/10.13039/501100011033.

}

}

\maketitle

\begin{abstract}
Federated Learning (FL) has emerged as a solution for distributed model training across decentralized, privacy-preserving devices, but the different energy capacities of participating devices (system heterogeneity) constrain real-world implementations. These energy limitations not only reduce model accuracy but also increase dropout rates, impacting on convergence in practical FL deployments. In this work, we propose LeanFed, an energy-aware FL framework designed to optimize client selection and training workloads on battery-constrained devices. LeanFed leverages adaptive data usage by dynamically adjusting the fraction of local data each device utilizes during training, thereby maximizing device participation across communication rounds while ensuring they do not run out of battery during the process. We rigorously evaluate LeanFed against traditional FedAvg on CIFAR-10 and CIFAR-100 datasets, simulating various levels of data heterogeneity and device participation rates. Results show that LeanFed consistently enhances model accuracy and stability, particularly in settings with high data heterogeneity and limited battery life, by mitigating client dropout and extending device availability. This approach demonstrates the potential of energy-efficient, privacy-preserving FL in real-world, large-scale applications, setting a foundation for robust and sustainable pervasive AI on resource-constrained networks.
\end{abstract}

\IEEEpeerreviewmaketitle

\section{Introduction}

Until recently, most machine learning (ML) solutions relied on centralized training, where data is transferred from local devices to powerful data centers. Although effective, this approach raises significant privacy and environmental concerns, largely due to the high energy consumption and carbon footprint associated with data transmission and intensive computations in large-scale data centers \cite{avgerinou2017trends, baliga2010green}. As the use of AI expands globally, the environmental impact of centralized frameworks is becoming unsustainable, given the substantial energy requirements of these centralized systems \cite{patterson2021carbon, strubell2020energy}. Recent studies indicate that transitioning from centralized to distributed computing architectures could reduce the total carbon footprint of ML operations, advancing more sustainable, energy-efficient AI \cite{ahvar2019estimating, guerra2023cost}.

In response, pervasive AI has emerged as a promising solution, enabling distributed learning directly on edge devices and bringing AI processing closer to data sources \cite{chen2021distributed, wang2023eidls, yuan2024distributed}.
By processing data locally, pervasive AI minimizes the need to transmit large volumes of raw data to centralized servers, thereby also reducing bandwidth requirements and addressing privacy concerns. This shift not only improves response times and network efficiency but also enables a more privacy-preserving, sustainable, and scalable approach to machine learning \cite{verbraeken2020survey}.

Nonetheless, despite these benefits, enabling truly pervasive AI solutions comes with its own set of challenges. Unlike cloud-based AI, which relies on high-performance centralized infrastructure, edge AI must leverage collaboration across multiple heterogeneous, resource-constrained devices operating in dynamic environments. These devices often have limited computational power, and energy resources, and can only access local data. Federated learning (FL) is a widely adopted distributed learning approach that moves training closer to data sources, minimizing the need for centralized processing \cite{Imteaj_survey_constraint_fl, Abdulrahman_motivation_fl_aiot}. In FL, each client/device trains its model locally on private data and shares only updated model parameters with a central server, which then aggregates these updates to refine a global model for distribution to all clients. This iterative process of local computation and centralized aggregation significantly reduces data transfer, addressing both privacy and energy efficiency.

Unlike centralized learning, the convergence rate in FL models is strongly influenced by data distribution and client participation in each training round \cite{li2019convergence, lu2024federated, li2020federated}. Particularly, when data is independent and identically distributed (iid) across devices, training on all devices may yield accuracy and convergence rates comparable to those in centralized solutions \cite{zhao2018federated, lian2017can}. In real-world applications, however, data is typically heterogeneously distributed (non-iid) among clients, and only a subset of devices participate in each training round. Therefore, convergence depends heavily on the number and specific selection of participating devices \cite{wang2020optimizing, zhang2021client, cho2020client}.

A larger device pool is generally associated with more data, which potentially accelerates convergence and increases model accuracy. However, to truly promote pervasive AI, FL must also be deployed on resource-constrained devices with varying, and often limited, battery capacities.

In this scenario, recent studies have shown that energy heterogeneity among devices can cause dropout and gradient divergence, impacting performance significantly \cite{arouj2022towards}. 
For instance, client selection strategies that disregard battery life can accelerate battery drainage, leading to temporary or prolonged device unavailability during training \cite{albelaihi2022green}. 
In this work, we argue that prioritizing devices' battery life is essential for enhancing global model performance in resource-limited FL scenarios. Particularly, we propose selecting a subset of each device's local data to extend its battery life and participation in the FL process. Although training on reduced data may yield less accurate local models temporarily, it significantly conserves battery life, enabling devices to participate across more training rounds. Our numerical experiments demonstrate that it is generally more advantageous for devices to engage in more rounds with fewer data samples than to participate in fewer rounds using their entire dataset. This strategy mitigates client dropout, preserves device energy, and ensures balanced contributions, thereby promoting robust convergence and enhancing the quality of the global model.

\section{Background and Motivation}
\label{sec:background}

Before presenting our energy-aware sampling strategy in Section~\ref{sec:energy_fl}, we first outline key concepts of federated learning, emphasizing centralized FL, the challenges posed by heterogeneous (non-iid) data distribution, and considerations related to local energy consumption.

In a typical centralized FL setup, a central server manages the learning process over a network of $E$ edge devices, where each of them owns a local dataset $D_e$, for $e=1, \ldots, E$. The objective of centralized FL is to minimize a weighted global loss function by aggregating information across the different devices while ensuring data privacy. Formally, this optimization process can be defined as
\begin{equation} 
\min_{w} f(w) = \sum_{e = 1}^E \frac{|D_e|}{|D|} F_e(w), 
\end{equation}
where  $F_e(w) = \mathbb{E}_{x_e \sim D_e}[f_e(w; x_e)]$ represents the local loss function computed by the $e$-th edge device on its local dataset $D_e$, and $|D_e|$ denotes the number of samples available to the $e$-th device. Here, $|D|$ quantifies the total number of samples across all devices.

This optimization typically is performed over multiple communication rounds. In each round \( r = 1, \ldots, R \), each device \( e = 1, \ldots, E \) initializes its local model \( w_e^{(r)} \) with the current global model \( w^{(r)} \), which has been broadcast by the server. Each device then performs \( L_e \) epochs of local training on its dataset \( D_e \). After completing the local updates, the server aggregates the models from each device to update the global model by computing the weighted average defined by
\begin{equation}
\label{eq:global_model_update}
w^{(r+1)} = \sum_{e=1}^E \frac{|D_e|}{|D|} w_e^{(r)}.
\end{equation}

In an ideal scenario where all devices have similar resources—such as balanced and similar data distributions, processing power, network stability, and battery life—the vanilla Federated Averaging (FedAvg) strategy~\cite{mcmahan2017communication} can yield optimal results. With independently and identically distributed (iid) data across devices, each local dataset closely approximates the global data distribution. This scenario minimizes the divergence between local and global models during aggregation, enabling the FL system to converge toward a solution comparable to that of a centralized model (depending on factors such as \( L_e, e = 1,..., E \) and network conditions).

In real-world applications, however, devices often exhibit significant heterogeneity in terms of data distribution, available resources, and connectivity. Specifically, the local data 
on each device may follow diverse distributions, with variations such as covariate shift, concept drift, or label distribution skew~\cite{kairouz2021advances, li2022federated, hsu2019measuring}. In this work, we focus on label heterogeneity, where label distributions vary across devices. For instance, in a location-dependent data collection scenario, certain classes may be overrepresented or underrepresented on specific devices, leading to imbalances that can worsen model convergence and degrade accuracy.

In addition to data heterogeneity, differences in device resources also affect the FL process. For instance, FedAvg assumes~\cite{mcmahan2017communication} that all devices perform the same number of local epochs (\( L = L_1 = \cdots = L_e \)). However, this assumption may not hold in real-world settings where devices vary in processing power, battery life, and network stability. Resource-constrained devices, or “stragglers,” can slow down FL by delaying communication rounds or providing less reliable updates. 

To address these challenges, adaptive methods like FedProx~\cite{li2020federated} enable devices to perform a variable number of local training steps, promoting more flexible participation based on resource availability. However, many existing approaches that aim to enable learning in resource-contained conditions primarily focus on the number of local epochs~\cite{li2020federated}, networking conditions~\cite{chen2021distributed}, and/or partial network updates~\cite{wang2024go}, which can overlook the broader implications of resource constraints. In parallel, we argue that adaptive sampling strategies~\cite{sreenivasan2023mini, albelaihi2022green} can dynamically select devices for each round based on criteria such as data representativeness or available energy. This dynamic selection helps balance the trade-offs between energy consumption, convergence speed, and model accuracy, ultimately optimizing FL performance in heterogeneous environments.

Building on the methods mentioned above, in our work we explore leveraging the variations in battery life and data availability across devices during training. As later shown in our results, it is crucial to ensure that all devices remain active until the last communication rounds, as their participation can significantly influence the convergence of the learning process and enhance the overall model performance.

\section{Energy-aware Federated Learning}
\label{sec:energy_fl}

We aim to explore strategies that enable distributed learning in energy-constrained devices, particularly focusing on scenarios where multiple devices may lack sufficient battery life to participate in all communication rounds $R$ pre-defined by the central server.  More formally, consider a set of devices $e = 1, \dots, E$, each with a fixed battery energy budget $B_e$.

During each communication round $r = 1, \dots, R$, each local training incurs an energy cost $\bar{b}_e$ for each device. 

This energy cost is assumed to be proportional to the device’s dataset size and is relatively stable across rounds. Specifically, for $e = 1, \dots, E$, we define $\bar{b}_e$ as:
$
\bar{b}_e = b_e^{(1)} \approx \cdots \approx b_e^{(R)}
$
where experimental observations support that, under fixed conditions, the (expected) energy consumption per round remains relatively stable across different rounds\footnote{Our experiments suggest that, for a given device and dataset, energy consumption per round is largely invariant. These results are not presented here due to space constraints.}.

Considering the scenario described above, employing traditional FedAvg, a device can be capable of participating in a maximum of
\begin{equation}
    {R}_e =  \frac{B_e}{\bar{b}_e L_e}  \quad\quad e = 1, ..., E
    \label{eq:max_rounds}
\end{equation}
communication rounds. 

If, for any given device, the pre-defined number of communication rounds $R$ is larger than the maximum number of rounds ${R}_e$, the device will run out of battery which will likely degrade the performance of the global model. This becomes especially problematic when $R$ is much larger than $R_e$ for a large number of devices (see Sec.~\ref{sec:results:learning_battery_constrained} for numerical results).

To ensure that $R_e \geq R$ for all devices, a simple yet effective approach is to reduce the cost of local learning by further reducing the amount of data used during local training at each device. Specifically, we propose that each device $e = 1, \dots, E$ uses only a percentage of its local data
\begin{equation}
    \eta_e(\lambda) =  \frac{1}{\lambda R }\frac{B_e}{\bar{b}_eL_e}  
    \label{eq:pct_data_to_use}
\end{equation}
where $\lambda \in [0,1]$ is defined by the central server. This fraction of data $\eta_e(\lambda)$ is inversely proportional to  $R$, $R_e$ and the device participation. By doing so, we ensure that each device’s participation capacity meets the maximum number of communication rounds $R$, thus preventing devices from running out of battery before the end of training.  We always enforce $\eta_e(\lambda) \in [0, 1]$, and hereafter, we will drop the dependence on $\lambda$ and define $\eta_e = \eta_e(\lambda)$.

Finally, when only a subset of devices participates per communication round, inserting the scaling factor $\lambda$ enables each device to use the maximum amount of local data possible while still preserving battery life for expected rounds. 

Our proposed framework is illustrated in Algorithm \ref{alg:lean_fed}. Setting $\eta_e = 1$ for all devices, our approach recovers the vanilla FedAvg method, where all devices use their entire dataset for each communication round.

\begin{algorithm}
    \caption{Energy-Aware Federated Learning (LeanFed)}
    \label{alg:lean_fed}
    
    \KwIn{Number of communication rounds \( R \), number of devices \( E \), number of local training epochs \( L_e \), device energy budgets \( B_1, \dots, B_E \), energy consumption per round \( \bar{b}_1, \dots, \bar{b}_E \)}

    \For{each communication round \( r = 1, \dots, R \)}{
        Server broadcasts the global model \( w^{(r)} \) to all devices\;
        
        \For{each device \( e = 1, \dots, E \) \textbf{in parallel}}{
            \If{\( B_e > 0 \)}{
                Device \( e \) initializes \( w_e^{(r)} \leftarrow w^{(r)} \)\;
                Device \( e \) estimates the percentage of data \(\eta_e\) to use according to  (\ref{eq:pct_data_to_use})\;
                
                \For{local epoch \( i = 1, \dots, L_e \)}{
                    \If{\( B_e > 0 \)}{
                        Device \( e \) trains on \( \eta_e \) percentage of its local data \( D_e \)\;
                        Update remaining energy \( B_e \leftarrow B_e - \bar{b}_e \cdot \eta_e  \)\;
                    }
                }
                
                \If{\( B_e > 0 \)}{
                    Device \( e \) sends updated model \( w_e^{(r)} \) to the server\;
                }
            }
        }
        
        Server aggregates the local models to update the global model \( w^{(r+1)} \) of all active devices according to Equation (\ref{eq:global_model_update})\;
    }
    
    \KwOut{Final global model \( w^{(R)} \)}
\end{algorithm}

\section{Numerical Evaluation}

\subsection{Experimental Setup}

\textbf{Datasets and baselines. }
In this section, we evaluate the impact of the energy-preserving strategy outlined above on both learning performance and energy consumption, comparing it to traditional implementations of FedAvg. 
We consider two standard benchmarks during our experiments: CIFAR--10 and CIFAR--100~\cite{krizhevsky2009learning_cifar} 
with various levels of data heterogeneity and participation rates. 
We simulate data heterogeneity across devices by sampling the ratio of data associated with each label from a Dirichlet distribution~\cite{hsu2019measuring}, which is controlled using a concentration parameter $\gamma \in \{0.5, 1.0\}$. 

For the i.i.d. scenario, we set $\gamma = 10^3$, which results in a near-homogeneous distribution across devices.
During evaluation, we use each dataset's full test set and record the highest test accuracy achieved during training, averaging over five independent executions of the experiment.  We compare our method, dubbed as {LeanFed}, with the vanilla FedAvg algorithm~\cite{mcmahan2017communication}.  We consider varying device participation rates ($\lambda \in \{80\%, 50\%, 20\%, 10\%\}$) to assess how reducing the number of participating devices affects training duration and energy consumption. Furthermore, for both methods, we assume that any device with remaining energy can participate in a training round, ensuring that energy constraints are respected throughout the learning process.

\textbf{Implementation details. }
We adopt a ResNet-18 as the deep architecture and train it from scratch using ADAM optimizer with a learning rate of $0.01$, and a weight decay of $10^{-4}$. Following prior works \cite{mcmahan2017communication}, we set the number of local epochs to $5$ and the batch size to $64$ throughout all experiments.
To simulate different battery resources at each device, we model $B_e$ as $B_e = (\alpha_e |D_e|/|D|)\cdot (\beta_e R)$, where $\alpha_e$, $\beta_e$ are drawn from a Gaussian distribution, i.e.,  $\alpha_e, \beta_e \sim \mathcal{N}(0.5, 0.5)$ and clipped to the range $ \alpha_e, \beta_e \in [0.1, 1]$.

The random variable $\beta_e$ controls the maximum number of communication rounds each device can perform with $\alpha_e$ percentage of its local data, simulating the proportionality of energy consumption relative to data size. We set   $b_e = |D_e| / |D|$. Finally, we used \textit{PyTorch} framework for implementation and executed all the experiments on a single NVIDIA GeForce RTX GPU 3090. We also make the simulation code available at:
\url{https://github.com/robertomatheuspp/leanFed}.

\subsection{Learning on Battery Constrained Devices}
\label{sec:results:learning_battery_constrained}

\begin{figure}[tb]
    \hspace*{0.2in}
    \scalebox{0.9}{
    \begin{subfigure}{0.7\textwidth}
        {\input{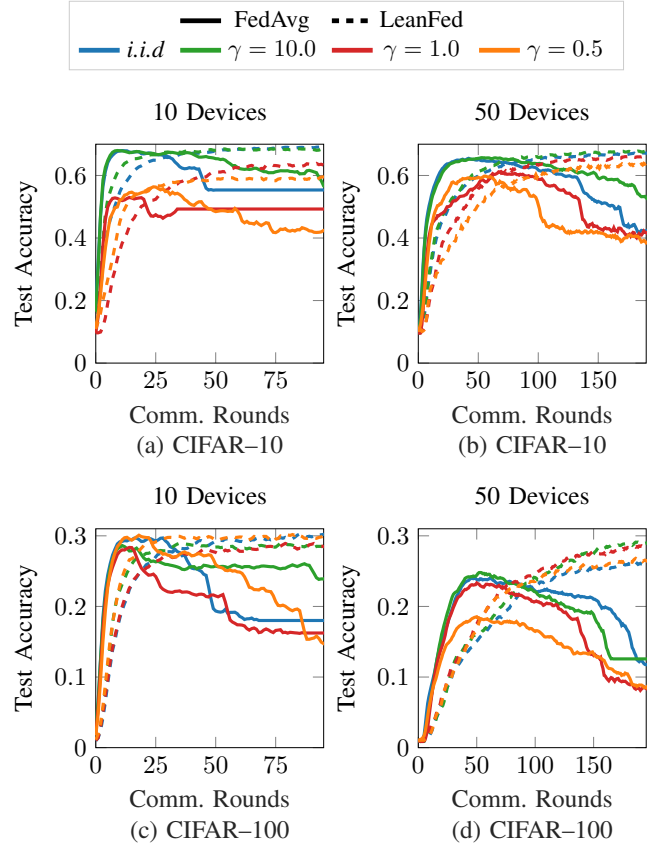}\vspace{1em}}
    \end{subfigure}}
    
     \scalebox{0.95}{\hspace*{-0.1in}
     \begin{subfigure}[b]{0.175\textwidth}
        \include{fig/fig1/fig1_top}
     \end{subfigure}
     }
     \hspace*{0.35in}
     \scalebox{0.95}{\begin{subfigure}[b]{0.175\textwidth}
        \include{fig/fig1/fig1_down}
     \end{subfigure}}
     
    \vspace{-1.5\baselineskip}
      \scalebox{0.95}{\hspace*{-0.1in}
     \begin{subfigure}[b]{0.175\textwidth}
        \include{fig/fig1/fig1c}
     \end{subfigure}
     }
     \hspace*{0.35in}
     \scalebox{0.95}{\begin{subfigure}[b]{0.175\textwidth}
        \include{fig/fig1/fig1d}
     \end{subfigure}}
      \vspace{-1.5\baselineskip}
      \caption{
       Test accuracy (y-axis) over communication rounds  (x-axis) for the baseline FedAvg (solid lines) and LeanFed (dashed lines) across various levels of data heterogeneity (indicated by color) and full device participation. Top row shows results for the CIFAR-10 dataset, while the bottom row presents results for the CIFAR-100. 
      }
    \label{fig:convergence_baseline_ours_100} 
    \vspace{-1\baselineskip}
\end{figure}

We start by comparing the convergence rate of our proposed method, LeanFed, against the baseline FedAvg on the CIFAR--10 and CIFAR--100 datasets.  Fig.~\ref{fig:convergence_baseline_ours_100} displays the test accuracy, averaged over five independent simulations, for both methods under varying data distribution settings (indicated by different colors in the plot).  
We consider two different scenarios, one with $10$ devices and another one with $50$ devices, training for $R=100$ and $R=200$ communication rounds, respectively. 
We notice that the test accuracy of LeanFed (dashed lines) consistently increases with the number of communication rounds, while the baseline (solid lines) reaches a peak accuracy before starting to degrade. During our experiments, we observed that this degradation occurs consistently regardless of the  level of data heterogeneity ($\gamma$), desired number of training rounds ($R$), number of devices ($E$), and datasets (CIFAR--10/100), suggesting that naively employing FedAvg may lead to undesired results in energy-constrained settings. We argue that this degradation in performance is a consequence of devices becoming inactive. Particularly, as the number of communication rounds increases a larger number of devices become inactive.
Consequently, the local model of the active ones tends to overfit to the local data, degrading the overall behavior of the global model\footnote{An alternative to avoid such degradation is for each device to only accept the global model if it does not degrade the performance on its local dataset. However, for simplicity, we opt to employ a vanilla FL approach where the global model always overwrites the local one.}.

\begin{table*}[ht]
\begin{center}
\caption{Average and variance of test accuracy of FedAvg and LeanFed considering a total of $50$ devices for varying levels of data heterogeneity and pre-defined communication rounds $R = 100$ and $200$. The highest accuracies are highlighted in bold.
}
\label{tab:results_diff_datasets}
\setlength\tabcolsep{7.9pt}
\hspace{-0.2cm}
\scalebox{0.95}{
\begin{tabular}{clcccccccccc} 
\toprule
\multirow{2}{*}{Dataset} & \multirow{2}{*}{Method}                                  
 & \multicolumn{2}{c}{$\gamma = 0.5$} & \multicolumn{2}{c}{$\gamma = 1.0$} & \multicolumn{2}{c}{\textit{i.i.d.}}  \\
& 
& \multicolumn{1}{c}{100 Rounds} & \multicolumn{1}{c}{200 Rounds}   &\multicolumn{1}{c}{100 Rounds } & \multicolumn{1}{c}{200 Rounds} &\multicolumn{1}{c}{100 Rounds} & \multicolumn{1}{c}{200  Rounds}             \\ 
\midrule
\multirow{6}{*}{CIFAR--10} 
& FedAvg & $52.90 \pm 2.67$ & $57.29 \pm 2.67$  & $58.33 \pm 2.45$  & $60.95 \pm 2.60$  & $63.11 \pm 0.32$  & $65.72 \pm 0.75$ \\
& FedAvg ($80\%$) & $52.34 \pm 2.51$ & $57.47 \pm 2.87$  & $57.56 \pm 2.35$  & $61.12 \pm 2.01$  & $63.25 \pm 1.08$ & $65.57 \pm  0.44$ \\
& FedAvg ($50\%$) & $55.30 \pm 2.89$ & $61.23 \pm 1.40$  & $59.99 \pm 2.29$  & $62.84 \pm 2.71$  & $65.19 \pm 0.99$ & $67.79 \pm  0.29$ \\
& FedAvg ($20\%$) & $59.41 \pm 2.42$ & $61.22 \pm 1.42$  & $\mathbf{62.87 \pm 0.87}$  & $64.12 \pm 0.96$  & $\mathbf{67.16 \pm 0.54}$ & $67.79 \pm  0.42$ \\
& FedAvg ($10\%$) & $57.40 \pm 1.49$ & $59.84 \pm 2.12$  & $61.81 \pm 0.89$  & $63.27 \pm 1.00$  & $66.81 \pm 0.64$ & $67.81 \pm  0.40$ \\
& \textbf{LeanFed} (\textbf{ours}) & $\mathbf{60.30 \pm 1.28}$ & $\mathbf{63.71 \pm 1.27}$  & $\mathbf{62.91 \pm 1.08}$  & $\mathbf{65.38 \pm 1.60}$  & $\mathbf{66.68 \pm 0.52}$  & $\mathbf{68.27 \pm 0.81}$ \\
\midrule
\multirow{6}{*}{CIFAR-100} 
& FedAvg & $17.82 \pm 0.99$ & $20.23 \pm 0.72$ & $20.09 \pm 1.41$ & $23.25 \pm 1.29$ & $21.34 \pm 1.70$ & $23.83 \pm 0.93$ \\
& FedAvg ($80\%$) & $18.17 \pm 0.69$ & $20.91 \pm 0.16$ & $19.79 \pm 1.58$ & $23.40 \pm 0.86$ & $21.54 \pm 1.21$ & $24.18 \pm 1.14$ \\
& FedAvg ($50\%$) & $20.45 \pm 1.13$ & $24.99 \pm 1.28$ & $21.86 \pm 1.27$ & $25.76 \pm 0.67$ & $22.59 \pm 1.30$ & $27.95 \pm 0.49$ \\
& FedAvg ($20\%$) & $\mathbf{25.44 \pm 1.25}$ & $26.08 \pm 1.07$ & $\mathbf{26.60 \pm 1.11}$ & $27.26 \pm 0.88$ & $27.38 \pm 0.55$ & $28.84 \pm 0.84$ \\
& FedAvg ($10\%$) & $24.07 \pm 1.49$ & $26.86 \pm 0.64$ & $\mathbf{26.48 \pm 1.28}$ & $27.70 \pm 0.55$ & $\mathbf{28.50 \pm 0.61}$  & $\mathbf{29.25 \pm 0.08}$ \\
& \textbf{LeanFed} (\textbf{ours}) & $\mathbf{25.20 \pm 1.73}$ & $\mathbf{28.32 \pm 1.05}$ & $\mathbf{26.02 \pm 2.64}$ & $\mathbf{30.51 \pm 1.24}$ & $26.85 \pm 2.23$ & $\mathbf{29.95 \pm 1.98}$ \\
\bottomrule
\end{tabular}}
\end{center}
\vspace{-3mm}
\end{table*}

Moreover, we also observe that the final accuracy achieved by LeanFed is consistently higher than the maximum peak values obtained by FedAvg. While also evident in homogeneous setup, this trend is more pronounced in highly heterogeneous scenarios, e.g., for $\gamma = 1.0$ (red) and  $\gamma = 0.5$ (orange) in Fig.~\ref{fig:convergence_baseline_ours_100}. In these settings, prolonging the participation of all the devices until the last communication round also seems to increase global accuracy.

To further explore the benefit of prolonging the device participation, we extend our analysis by comparing LeanFed with the FedAvg baseline across varying device participation rates on both the CIFAR-10 and CIFAR-100 datasets.  Table \ref{tab:results_diff_datasets} displays the highest test accuracy achieved by the global model (averaged over five independent simulations),  showing that LeanFed consistently outperforms all baseline configurations across both datasets.  As expected, increasing data heterogeneity among devices (lower values of $\gamma$) leads to a general degradation in performance across all methods, while increasing the number of communication rounds improves the overall test accuracy.

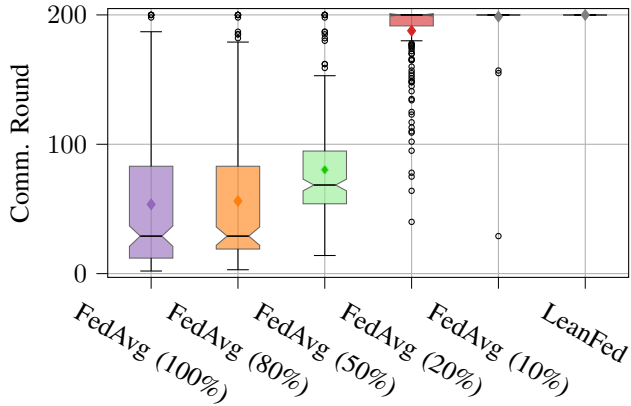
\begin{figure}[tb]
    \hspace*{0.05in}

     \begin{subfigure}[b]{0.38\textwidth}
\begin{tikzpicture}

\definecolor{crimson2143940}{RGB}{214,39,40}
\definecolor{darkblue}{RGB}{0,0,139}
\definecolor{darkgray176}{RGB}{176,176,176}
\definecolor{darkorange25512714}{RGB}{255,127,14}
\definecolor{darkturquoise23190207}{RGB}{23,190,207}
\definecolor{gray127}{RGB}{127,127,127}
\definecolor{lightgray204}{RGB}{204,204,204}
\definecolor{lightgreen}{RGB}{144,238,144}
\definecolor{darkgreen}{RGB}{31,198,0}
\definecolor{mediumpurple148103189}{RGB}{148,103,189}

\begin{axis}[
width=1\linewidth,
height=1.4in,
at={(0in,1.9in)},
scale only axis,
legend style={fill opacity=0.8, draw opacity=1, text opacity=1, draw=lightgray204},
tick align=outside,
tick pos=left,
x grid style={darkgray176},
xmajorgrids,
xmin=0.5, xmax=6.5,
xtick style={color=black},
ytick={0, 100, 200},
xtick={1,2,3,4,5,6},
xticklabel style={rotate=-30.0},
xticklabels={FedAvg (100\%),FedAvg (80\%),FedAvg (50\%),FedAvg (20\%),FedAvg (10\%),LeanFed},
y grid style={darkgray176},
ylabel={Comm. Round},
ymajorgrids,
ymin=-2.9, ymax=204.9,
ytick style={color=black}
]
\addplot [draw=mediumpurple148103189, fill=mediumpurple148103189, forget plot, mark=diamond*, only marks]
table{%
x  y
1 53.57
};
\addplot [draw=darkorange25512714, fill=darkorange25512714, forget plot, mark=diamond*, only marks]
table{%
x  y
2 56.065
};
\addplot [draw=lightgreen, fill=darkgreen, forget plot, mark=diamond*, only marks]
table{%
x  y
3 80.285
};
\addplot [draw=crimson2143940, fill=crimson2143940, forget plot, mark=diamond*, only marks]
table{%
x  y
4 187.75
};
\addplot [draw=gray, fill=gray, forget plot, mark=diamond*, only marks]
table{%
x  y
5 198.705
6 200
};
\path [draw=black, fill=mediumpurple148103189, opacity=0.6]
(axis cs:0.75,12)
--(axis cs:1.25,12)
--(axis cs:1.25,21.1178807101136)
--(axis cs:1.125,29)
--(axis cs:1.25,36.8821192898864)
--(axis cs:1.25,83)
--(axis cs:0.75,83)
--(axis cs:0.75,36.8821192898864)
--(axis cs:0.875,29)
--(axis cs:0.75,21.1178807101136)
--(axis cs:0.75,12)
--cycle;
\addplot [black, forget plot]
table {%
1 12
1 2
};
\addplot [black, forget plot]
table {%
1 83
1 187
};
\addplot [black, forget plot]
table {%
0.875 2
1.125 2
};
\addplot [black, forget plot]
table {%
0.875 187
1.125 187
};
\addplot [black, mark=o, mark size=1, mark options={solid,fill opacity=0}, only marks, forget plot]
table {%
1 200
1 198
1 200
1 200
1 200
1 200
};
\path [draw=black, fill=darkorange25512714, opacity=0.6]
(axis cs:1.75,19)
--(axis cs:2.25,19)
--(axis cs:2.25,21.8949910626376)
--(axis cs:2.125,29)
--(axis cs:2.25,36.1050089373624)
--(axis cs:2.25,83)
--(axis cs:1.75,83)
--(axis cs:1.75,36.1050089373624)
--(axis cs:1.875,29)
--(axis cs:1.75,21.8949910626376)
--(axis cs:1.75,19)
--cycle;
\addplot [black, forget plot]
table {%
2 19
2 3
};
\addplot [black, forget plot]
table {%
2 83
2 179
};
\addplot [black, forget plot]
table {%
1.875 3
2.125 3
};
\addplot [black, forget plot]
table {%
1.875 179
2.125 179
};
\addplot [black, mark=o, mark size=1, mark options={solid,fill opacity=0}, only marks, forget plot]
table {%
2 187
2 200
2 185
2 182
2 198
2 200
2 185
2 200
2 200
2 200
};
\path [draw=black, fill=lightgreen, opacity=0.6]
(axis cs:2.75,54)
--(axis cs:3.25,54)
--(axis cs:3.25,63.9761075906638)
--(axis cs:3.125,68.5)
--(axis cs:3.25,73.0238924093362)
--(axis cs:3.25,94.75)
--(axis cs:2.75,94.75)
--(axis cs:2.75,73.0238924093362)
--(axis cs:2.875,68.5)
--(axis cs:2.75,63.9761075906638)
--(axis cs:2.75,54)
--cycle;
\addplot [black, forget plot]
table {%
3 54
3 14
};
\addplot [black, forget plot]
table {%
3 94.75
3 153
};
\addplot [black, forget plot]
table {%
2.875 14
3.125 14
};
\addplot [black, forget plot]
table {%
2.875 153
3.125 153
};
\addplot [black, mark=o, mark size=1, mark options={solid,fill opacity=0}, only marks, forget plot]
table {%
3 187
3 200
3 185
3 182
3 162
3 198
3 159
3 200
3 180
3 187
3 200
3 200
3 200
3 162
};
\path [draw=black, fill=crimson2143940, opacity=0.6]
(axis cs:3.75,191.5)
--(axis cs:4.25,191.5)
--(axis cs:4.25,199.056366000507)
--(axis cs:4.125,200)
--(axis cs:4.25,200.943633999493)
--(axis cs:4.25,200)
--(axis cs:3.75,200)
--(axis cs:3.75,200.943633999493)
--(axis cs:3.875,200)
--(axis cs:3.75,199.056366000507)
--(axis cs:3.75,191.5)
--cycle;
\addplot [black, forget plot]
table {%
4 191.5
4 180
};
\addplot [black, forget plot]
table {%
4 200
4 200
};
\addplot [black, forget plot]
table {%
3.875 180
4.125 180
};
\addplot [black, forget plot]
table {%
3.875 200
4.125 200
};
\addplot [black, mark=o, mark size=1, mark options={solid,fill opacity=0}, only marks, forget plot]
table {%
4 102
4 177
4 172
4 178
4 134
4 148
4 135
4 174
4 155
4 113
4 78
4 170
4 95
4 151
4 175
4 166
4 176
4 149
4 117
4 75
4 145
4 153
4 123
4 165
4 165
4 177
4 170
4 148
4 141
4 40
4 110
4 125
4 160
4 157
4 109
4 64
};
\path [draw=black, fill=gray127, opacity=0.6]
(axis cs:4.75,200)
--(axis cs:5.25,200)
--(axis cs:5.25,200)
--(axis cs:5.125,200)
--(axis cs:5.25,200)
--(axis cs:5.25,200)
--(axis cs:4.75,200)
--(axis cs:4.75,200)
--(axis cs:4.875,200)
--(axis cs:4.75,200)
--(axis cs:4.75,200)
--cycle;
\addplot [black, forget plot]
table {%
5 200
5 200
};
\addplot [black, forget plot]
table {%
5 200
5 200
};
\addplot [black, forget plot]
table {%
4.875 200
5.125 200
};
\addplot [black, forget plot]
table {%
4.875 200
5.125 200
};
\addplot [black, mark=o, mark size=1, mark options={solid,fill opacity=0}, only marks, forget plot]
table {%
5 157
5 29
5 155
};
\path [draw=black, fill=darkturquoise23190207, opacity=0.6]
(axis cs:5.75,200)
--(axis cs:6.25,200)
--(axis cs:6.25,200)
--(axis cs:6.125,200)
--(axis cs:6.25,200)
--(axis cs:6.25,200)
--(axis cs:5.75,200)
--(axis cs:5.75,200)
--(axis cs:5.875,200)
--(axis cs:5.75,200)
--(axis cs:5.75,200)
--cycle;
\addplot [black, forget plot]
table {%
6 200
6 200
};
\addplot [black, forget plot]
table {%
6 200
6 200
};
\addplot [black, forget plot]
table {%
5.875 200
6.125 200
};
\addplot [black, forget plot]
table {%
5.875 200
6.125 200
};
\addplot [semithick, black, forget plot]
table {%
0.875 29
1.125 29
};
\addplot [semithick, black, forget plot]
table {%
1.875 29
2.125 29
};
\addplot [semithick, black, forget plot]
table {%
2.875 68.5
3.125 68.5
};
\addplot [semithick, black, forget plot]
table {%
3.875 200
4.125 200
};
\addplot [semithick, black, forget plot]
table {%
4.875 200
5.125 200
};
\addplot [semithick, black, forget plot]
table {%
5.875 200
6.125 200
};
\end{axis}
\end{tikzpicture}
     \end{subfigure}
      \vspace{-1.5\baselineskip}
      \caption{
       Number of communication rounds after which devices become inactive due to drained battery, considering $50$ devices, $R = 200$ and $\gamma = 0.5$. 
       Each boxplot corresponds to a different method and/or participation method.
           \vspace{-1\baselineskip}
      }
          \vspace{-1\baselineskip}
    \label{fig:results:boxplots_baseline_ours_100} 
\end{figure}

The above results suggest that in energy-constrained federated learning, a lower participation rate can be advantageous, as only selected devices expend battery resources during local training. 
This allows the remaining devices to conserve energy, potentially extending their active participation across more training rounds. 
However, if the participation rate is too low (e.g., $10\%$ in Table~\ref{tab:results_diff_datasets}), the global model may lack sufficient representation, leading to degraded performance. Conversely, if the participation rate is too high (e.g., $80\%$ or $50\%$), many devices may exhaust their batteries in the early training stages, reducing their availability for later rounds.
Figure~\ref{fig:results:boxplots_baseline_ours_100} illustrates this behavior through boxplots showing the number of devices that run out of battery across communication rounds under fixed $R = 200$,  $\gamma = 0.5$ and number of devices $E = 50$.  Bars near the bottom represent scenarios where a high number of devices become inactive early, while higher-positioned bars suggest better device availability throughout previous rounds. In the first three scenarios—FedAvg baseline with $100\%$, $80\%$ and $50\%$ participation rates—the majority of devices become inactive early in training. In contrast, when considering FedAvg at $20\%$ and $10\%$ participation rates, as well as LeanFed, the boxplots are closer to the maximum number of communication rounds, indicating prolonged device participation.

\begin{figure}[tb]
    \hspace*{0.05in}
    \scalebox{0.9}{
    \begin{subfigure}{0.7\textwidth}
        {\input{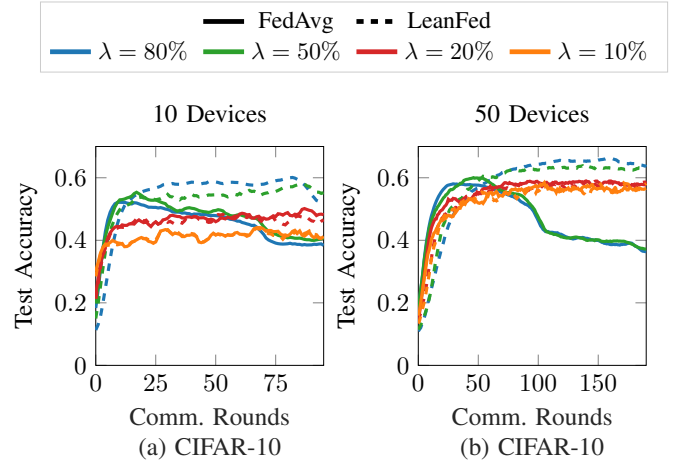}\vspace{1em}}
    \end{subfigure}}
    
     \scalebox{0.95}{\hspace*{-0.1in}
     \begin{subfigure}[b]{0.175\textwidth}
        \include{fig/fig3/fig3a}
     \end{subfigure}
     }
     \hspace*{0.35in}
     \scalebox{0.95}{\begin{subfigure}[b]{0.175\textwidth}
        \include{fig/fig3/fig3b}
     \end{subfigure}}

      \vspace{-1.5\baselineskip}
      \caption{Test accuracy of the FedAvg (solid line) and LeaFed (dashed line) for fixed $\gamma=0.5$ on CIFAR-10 (top) 
      and different participation rates (different colors). }
    \label{fig:result:diff_participation} 
    \vspace{-1\baselineskip}
\end{figure}

Finally, building on the observations above, we further investigate the behavior of LeanFed under partial device participation. Figure~\ref{fig:result:diff_participation} presents results similar to those in Fig.~\ref{fig:convergence_baseline_ours_100} but focuses on a fixed heterogeneous data distribution ($\gamma = 0.5$) and varying device participation rates, $\lambda \in \{ 80\%, 50\%, 20\%, 10\%\}$. In low participation settings ($\lambda \in \{10\%, 20\%\}$), where most devices remain active throughout training, LeanFed performs comparably to (or slightly better than) FedAvg. 
This is aligned with the results from (\ref{eq:pct_data_to_use}), which indicates that as $\lambda$ approaches zero, $\eta_e(\lambda)$ approaches one, effectively making LeanFed equivalent to FedAvg\footnote{Using low participation rates, e.g. $\lambda = 10\%$ does not grantee that all the energy of the devices have been consumed. Therefore, many of the local devices could have potentially participated in more communication rounds. }.

A more interesting observation happens when the participation rates increase. In this setting, LeanFed consistently outperforms both FedAvg and LeanFed under full device participation. Specifically, with $\lambda = 80\%$, LeanFed achieves peak test accuracies of $61.26~(\pm 1.44)$ for $E = 10$ and $66.63~(\pm 1.23)$ for $E = 50$, surpassing the accuracies presented in Table~\ref{tab:results_diff_datasets} which are obtained under full device participation. We hypothesize that this is due to a higher proportion of local data used per device when participation is partial, leading to enhanced model training. However, further investigation is needed to fully understand this effect, which we leave as future work.

\subsection{Limitations}

While LeanFed demonstrates significant advantages in scenarios where energy-constrained devices must participate in federated learning with limited communication opportunities, we recon that it may suffer limitations. Specifically, LeanFed is most effective when the number of communication rounds required for convergence is relatively high, and devices have insufficient battery levels to participate in every round. However, these benefits may become limited or almost negligible in scenarios where devices have large battery capacity or when the pre-defined number of communication rounds is significantly larger the actual number of rounds required to achieve convergence. We also highlight that LeanFed relies on accurate battery level estimations, which may not always be available in real-world conditions, potentially leading to suboptimal scheduling decisions. We leave such evaluations to future work.

\section{Conclusion}

In this paper, we have presented a simple and efficient approach to federated learning in energy-constrained environments, where device battery limitations can hinder long-term participation and compromise model performance. By adaptively adjusting the amount of data used during local training by each device,  our proposed LeanFed method ensures extended device availability across communication rounds without sacrificing model accuracy. Our experimental results  demonstrated that LeanFed consistently outperforms traditional FedAvg, especially in scenarios with high data heterogeneity and limited battery. Future works can extend this to consider dynamic energy adjustments, incorporating real-time device constraints, and exploring LeanFed's application across diverse federated learning scenarios and real-world deployments.

\bibliographystyle{IEEEtran}
\bibliography{./bibliography.bib}

\end{document}